# An Improved Model Ensembled of Different Hyper-parameter Tuned Machine Learning Algorithms for Fetal Health Prediction

**Md. Simul Hasan Talukder [1], Sharmin Akter[2]**

[1] Department of Electrical and Electronic Engineering, Rajshahi University of Engineering and Technology, Bangladesh; simulhasantalukder@gmail.com

[2] Department of Biomedical Engineering, Jashore University of Science and Technology, Bangladesh; sharmintalukder120@gmail.com

**Corresponding Author**: Md. Simul Hasan Talukder; Email: simulhasantalukder@gmail.com

**Abstract:** Fetal health is a critical concern during pregnancy as it can impact the well-being of both the mother and the baby. Regular monitoring and timely interventions are necessary to ensure the best possible outcomes. While there are various methods to monitor fetal health in the mother's womb, the use of artificial intelligence (AI) can improve the accuracy, efficiency, and speed of diagnosis. In this study, we propose a robust ensemble model called ensemble of tuned Support Vector Machine and ExtraTrees (ETSE) for predicting fetal health. Initially, we employed various data preprocessing techniques such as outlier rejection, missing value imputation, data standardization, and data sampling. Then, seven machine learning (ML) classifiers including Support Vector Machine (SVM), XGBoost (XGB), Light Gradient Boosting Machine (LGBM), Decision Tree (DT), Random Forest (RF), ExtraTrees (ET), and K-Neighbors were implemented. These models were evaluated and then optimized by hyperparameter tuning using the grid search technique. Finally, we analyzed the performance of our proposed ETSE model. The performance analysis of each model revealed that our proposed ETSE model outperformed the other models with 100% precision, 100% recall, 100% F1-score, and 99.66% accuracy. This indicates that the ETSE model can effectively predict fetal health, which can aid in timely interventions and improve outcomes for both the mother and the baby.

## 1. Introduction

Pregnancy is a joyous and memorable time for women, but it is also a period that requires special care for both the mother and fetus. Abnormal fetal growth poses significant risks to maternal health and mortality, with approximately 810 women dying every day during pregnancy according to WHO [1]. The maternal mortality ratio is higher in underdeveloped countries, highlighting the crucial need for fetal health monitoring. Doctors recommend regular tests to monitor the condition of the fetus, with Cardiotocography (CTG) being a widely used technique for assessing fetal and maternal health [2]. CTG continuously records the fetal heart rate (FHR) and uterine contractions (UC) via an ultrasound transducer located on the mother's abdomen [3]. The CTG data includes 21 attributes that obstetricians use to determine fetal health status, enabling early detection and treatment of fetal distress before it becomes severe [4,5]. The emergence of artificial intelligence has facilitated further advancements in fetal health monitoring [6-11]. However, accurately predicting fetal health conditions remains a major challenge in the field of machine learning.

In our work, we utilized the UCI CTG dataset to improve the accuracy of fetal health prediction. We proposed a highly accurate ensemble of tuned SVM and ExtraTrees (ETSE) model that outperforms existing literature. We applied various data preprocessing techniques such as outlier rejection, missing value imputation, data standardization, and data sampling to enhance the quality of the data fed into the models. Furthermore, we implemented hyper-parameter tuning using the Grid Search Technique to improve the accuracy of each conventional classifier used in fetal health prediction. Finally, we ensembled different models to create a benchmark model for accurate fetal health prediction. Our proposed approach demonstrates significant improvements in predicting fetal health conditions, indicating the potential of artificial intelligence in improving maternal and fetal health outcomes.

The paper is structured as follows: Section 1 provides an overview of the study, including the purpose and significance of the research. Section 2 reviews related works in the field and identifies gaps addressed by our study. Section 3 describes the materials and methods used in our research, including data sources, study design, and statistical analysis. Section 4 presents the results of the study in a clear and concise manner. Finally, Section 5 offers a comprehensive discussion of the results, including a critical evaluation of the study, its implications for the field, and recommendations for future research.

## 2. Related Work

Artificial intelligence (AI) is currently one of the most popular topics in healthcare automation, with extensive use in disease detection, diagnosis, and medical treatment. A review of the literature shows that numerous studies have been conducted on the prediction of various diseases, such as diabetes, heart disease, fetal health disorders, and many more. In a study by Amzad Hossen et al. [12], the performance of three supervised machine learning models, namely Random Forest (RF), Decision Tree (DT), and Logistic Regression (LR), was investigated in heart disease analysis and prediction. The results showed that logistic regression achieved the highest accuracy score of 92.10%. It is noted that this study did not explore other forms of conventionally supervised machine learning. However, Abdul Saboor et al. [13] used nine classifiers of machine learning including AdaBoost (AB), LR (Logistic regression), ET, MNB(Multinomial Naive Bayes), CART (Classification and Regression Trees), SVM, LDA (Linear Discriminant Analysis), RF, and XGB to predict human heart diseases. The experimental results revealed that the accuracy of the prediction classifiers improved with hyperparameter tuning and achieved notable results with data standardization and machine learning hyperparameter tuning. The authors also indicated that they are attempting to apply AI in fetal health disease in a similar manner to that of heart disease prediction. One such researcher is Sundar.C [14], who calculated precision, recall, and F-score for the commonly used unsupervised clustering method k-means clustering in the prediction of fetal health diseases. However, the performance of this method was not satisfactory. To improve upon these results, Sundar.C. et al. [3] implemented a model-based CTG data classification system using a supervised artificial neural network (ANN), which yielded significantly improved performance. The values of precision, recall, and F1 score all showed notable improvements. The authors Ocak and Ertunc [15] used adaptive neuro-fuzzy inference techniques to classify CTGs (ANFIS), while Ocak developed a classification algorithm based on support vector machine (SVM) and genetic algorithms (GA) [16]. Muhammad Arif et al. [4] proposed a Random Forest classifier to distinguish between normal, suspicious, and pathological patterns. The results showed that the random forest classifier successfully identified these patterns, achieving an overall classification accuracy of 93.6 percent. Similarly, Tomáš Peterek et al. [17] employed the Random

Forest method and achieved exceptional performance, achieving 94.69 percent accuracy in classifying the data. It is important to note that the author only utilized one supervised machine learning method and did not explore data processing, hyperparameter tuning, or other machine learning models. On the other hand, Abolfazl Mehbodniya et al. [18] applied four supervised machine learning models, namely support vector machine, random forest (RF), multi-layer perceptron, and K-nearest neighbours, to predict the health state of the fetus. The RF algorithm achieved an accuracy of 94.5% on the CTG dataset. However, the accuracy is still considered low. To improve upon this, Nabillah Rahmayanti et al. [19] emphasized the importance of outlier removal, multicollinearity removal using VIF, data balancing, data scaling, and standardization. The authors applied seven algorithms, namely Artificial Neural Network, Long-Short Term Memory (LSTM), XG-Boost (XGB), Support Vector Machine (SVM), Neighbour (KNN), Light GBM (LGBM), and Random Forest (RF), to predict fetal health and compared their performance. The LGBM algorithm performed well across all seven scenarios. Recently, Md Takbir Alam et al. [20] conducted an analysis of multiple machine learning models, including random forest (RF), logistic regression, decision tree (DT), support vector classifier, voting classifier, and K-nearest neighbor, to classify fetal health using the CTG dataset. The results of this study showed that the Random Forest classifier achieved the highest accuracy of 97.51%, surpassing the accuracy achieved by previous studies. A summary of the performance comparison of different fetal health classification study is provided in Table 1.

**Table 1**. Summary of findings from the litteraure review.

| **Ref.** | **Models** | **Dataset** | **Hyper parameters tuning** | **Ensemble Learning** | **Metric** | | |
|---|---|---|---|---|---|---|---|
| [14] | K-Means Clustering | Own (Normal, Suspect, Pathological) | No | No | F1-Score (Normal) 0.4575 | F1-Score (Suspect) 0.1776 | F1-Score (Pathological) 0.1403 |
| [3] | ANN | Own (Normal, Suspect, Pathological) | No | No | F1-Score (Normal) 0.9784 | F1-Score (Suspect) 0.4514 | F1-Score (Pathological) 0.9724 |
| [15] | ANFIS | Own (Normal, Suspect, Pathological) | No | No | Accuracy= 91.6%. | | |
| [16] | SVM and GA | Own (Normal and Abnormal) | No | No | Accuracy= 98% | | |

| | | | | | |
|---|---|---|---|---|---|
| [4] | RF | Own (Normal, Suspect, Pathological) | No | No | Accuracy= 93.6 |
| [17] | RF | Own (Normal, Suspect, Pathological) | No | No | Accuracy= 94.69 |
| [18] | RF | Own (Normal, Suspect, Pathological) | No | No | Accuracy= 94.5% |
| [20] | RF | **UCI Public Dataset (Normal, Suspect, Pathological)** | No | No | **Accuracy= 97.51%** |

According to the literature reviewed, a variety of advanced machine learning and data preprocessing techniques have been utilized for fetal health detection. However, most of the researchers have used their own datasets. In contrast, the authors of reference [20] used the publicly available UCI dataset and achieved an accuracy of 97.51%. However, it has been observed that hyperparameter techniques and ensemble learning have not yet been extensively explored.

In this study, we propose an ETSE model that applies hyperparameter tuning to different supervised machine learning models and analyzes the ensemble effect of these models. ExtraTrees classifiers are used in our work, which have not been used before on this dataset. The results demonstrate that the ensemble of tuned ET and SVM (ETSE) provides the best performance. Our work makes several important contributions to the field of fetal health classification using machine learning. Firstly, we propose a novel ETSE model that achieves high accuracy in fetal health classification. Secondly, we introduce various data preprocessing techniques, such as outlier rejection, missing value imputation, data standardization, and data sampling, to improve the quality of the data used in our analysis. Thirdly, we use Grid Search Techniques for Hyperparameters tuning to ensure the best performance of our models. Fourthly, we introduce the ExtraTrees model, which has not been used before on this dataset. Finally, we demonstrate the effectiveness of ensemble learning in improving the overall performance of our model. Overall, our work contributes to the growing body of literature on machine learning-based approaches to fetal health classification.

## 2. Materials and Methodology

Our work provides an efficient prediction of fetal health condition while in utero, which is critical for the well-being of both the mother and the unborn child. The methods used in this study are thoroughly described in the following sections.

### 3.1. Dataset

Our experiment utilized the publicly available cardiotocography (CTG) dataset from the UCI Machine Learning repository [21]. This dataset contains 2126 records of features extracted from cardiotocograph exams, which were classified into three categories based on the assessment of three expert obstetricians using 23 different attributes. The attributes used in the measurement of fetal heart rate (FHR) and uterine contractions (UC) on CTG are listed in Table 2, and the histogram of the dataset is shown in Figure 2. The dataset is imbalanced, with 1655 records classified as normal, 295 as suspect, and 176 as pathological, as illustrated in Figure 1.

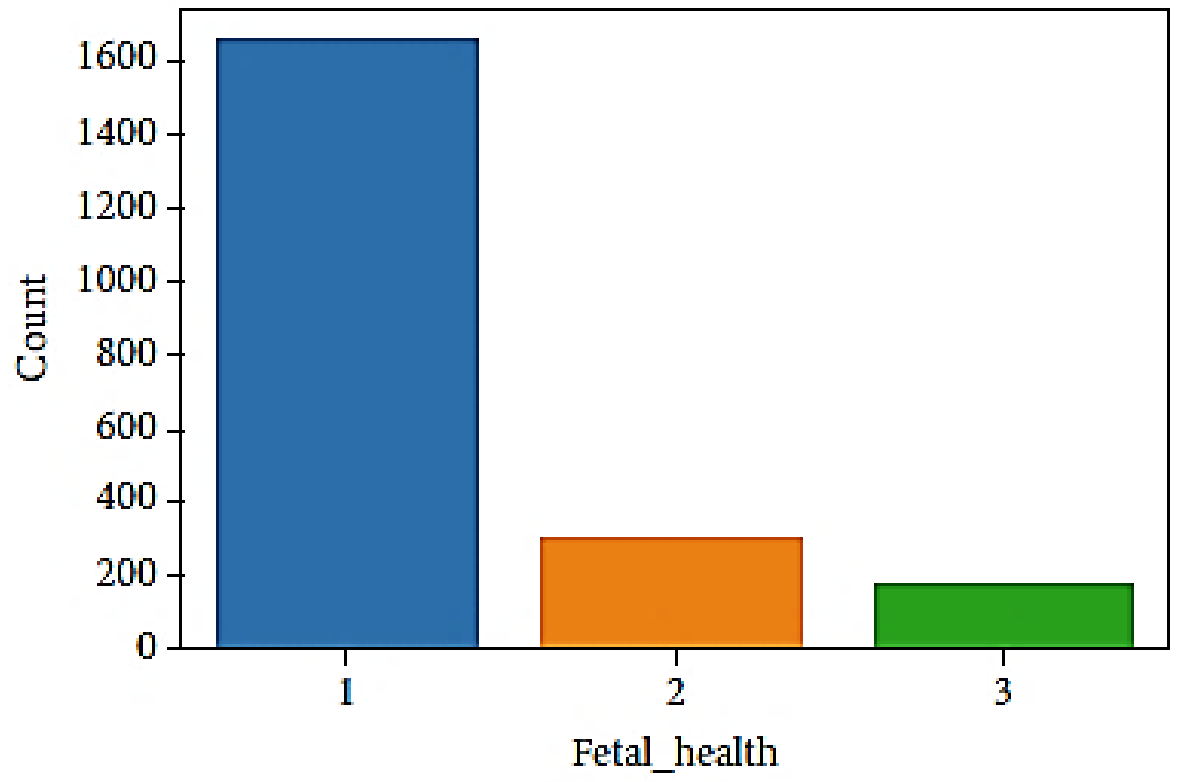

**Figure 1.** Data distribution in each class.

**Table 2**. CTG dataset atributes used in our model.

| Variable Symbol | Variable Description | Range |
|---|---|---|
| LB | FHR baseline (beats per minute) | (160–106) |
| AC | Accelerations per second | (0–0.019) |
| FM | Fetal movement (number of fetal movement) | (0–0.481) |
| UC | Uterine contractions (number of uterine contractions per second) | (0–0.015) |
| DL | Light decelerations (number of light decelerations per second) | (0–0.015) |
| DS | Severe decelerations (number of severe decelerations per second) | (0–0.001) |
| DP | Prolonged decelerations (number of prolonged decelerations per second) | (0–0.005) |
| ASTV | Abnormal short-term variability (percentage of time with abnormal short-term variability) | (12–87) |
| MSTV | Mean value of short-term variability | (7–0.2) |
| ALTV | Percentage of time with abnormal long-term variability | (0–91) |
| MLTV | Mean value of long-term variability | (0–50.7) |
| Width | Width of FHR histogram | (3–180) |
| Min | Minimum of FHR histogram | (50–159) |
| Max | Maximum of FHR histogram | (122–238) |
| Nmax | Histogram peaks | (0–18) |
| Nzeros | Histogram zeros | (0–10) |
| Mode | Histogram mode | (60–187) |
| Mean | Histogram means | (73–182) |
| Median | Histogram median | (77–186) |
| Variance | Histogram variance | (0–269) |
| Tendency | Histogram tendency | 1 |
| CLASS | FHR pattern class code (1 to 10) | (1-10) |
| NSP | Fetal health (Fetal state class code, N=normal, S=Suspected, P=Pathological) | (1-Normal, 2-Needs Reassurance, 3-Pathological) |

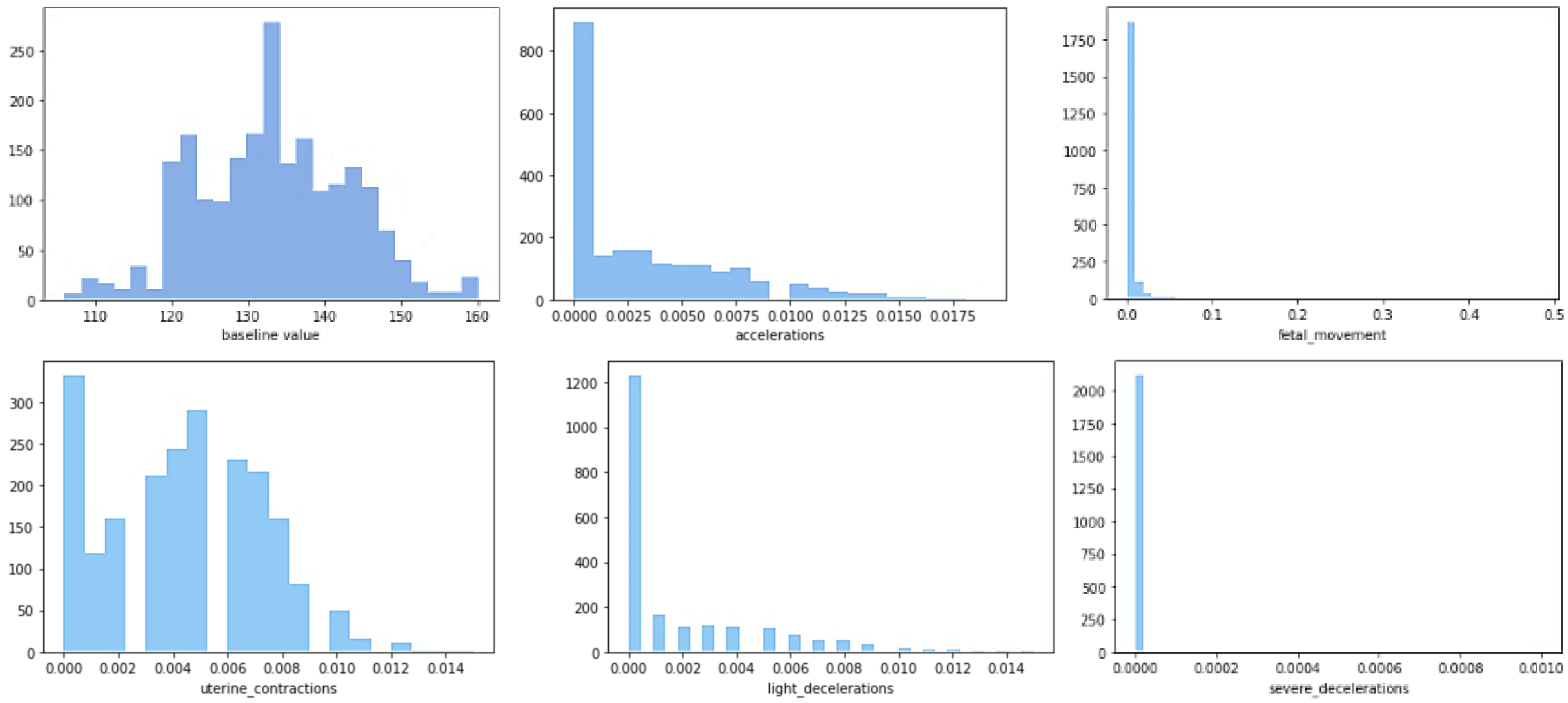

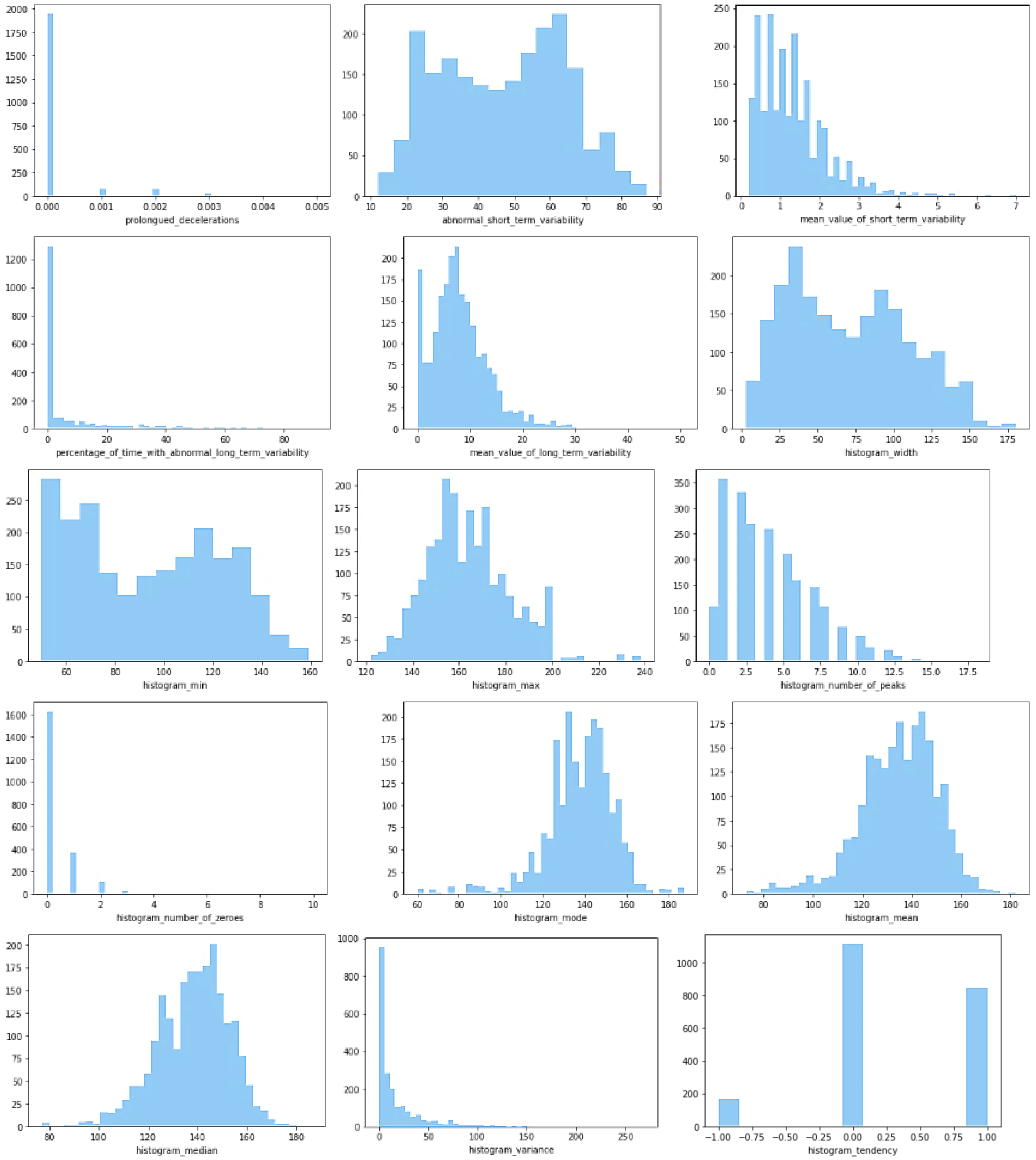

**Figure 2.** Histrograms of the Dataset.

### 3.2. Data Pre-processing

In order to improve the performance of our model, data processing is a crucial step, particularly given that the real-time dataset we are using contains missing values and noisy data. In our work, we conducted a check for missing data, but none was found. To deal with outliers, we used standard deviation to remove them. However, as our dataset is imbalanced, the minority class may be ignored by the model. To address this issue, we applied the Random Over Sampler (ROS) [22], which is a technique that duplicates the data of the minority class to balance the dataset. This approach helps to reduce overfitting on skewed classes and improve the overall performance of the model. Additionally, we used standardization (S) to rescale the distributions and achieve zero mean and one standard deviation. This process reduces the skewness of the data distribution, as shown in Eq. (1).

$$S(x) = \frac{x-\bar{x}}{\sigma} \tag{1}$$

Where, $X$ refers to the n-dimensional instances of the feature vector. As a result of standardizing, it results in a more stable model that is less influenced by variables, fits faster, and performs more consistently. Therefore, scaling is a crucial aspect of data pre-processing [23]. Once the data was standardized, we proceeded to split the dataset into training and testing sets, using a ratio of 70% for training and 30% for testing.

### 3.3. Grid Search and Hyper-parameter tunning

Hyperparameter tuning is a critical aspect of any machine learning pipeline, as the values of the model parameters must be set before training. Poorly chosen hyperparameter values can lead to incorrect results and poor model performance [24]. Therefore, finding optimal values in multidimensional space is crucial. The most common method for learning hyperparameter tuning is grid search (GS) [25]. In this method, hyperparameter domains are divided into discrete grids, and a combination of values from these grids is used to calculate model performance. The grid points with the highest performance are considered the most optimized combinations of hyperparameter values. The 'GridSearchCV' function in the sklearn library [26] can be used to apply the grid search algorithm and find the optimal hyperparameters. Grid search is widely suggested as a hyperparameter optimization method due to its easy implementation, reliability, and low dimensionality [27]. In our work, we performed grid search and hyperparameter tuning on all the fundamental models. This approach ensured that our model was optimized and had the best possible performance.

### 3.4. Ensemble Learning

Ensemble learning is a powerful technique that combines individual machine learning models (base-predictors) to create a superior model that addresses the shortcomings of the base-predictors, making it the most efficient way to improve the performance of traditional machine learning [28]. The voting classifier is an ensemble classifier that integrates different models and predicts through majority voting [29]. The output of these base predictors is combined using some combination rules, such as majority voting, minimum or maximum probabilities, or probabilistic product [28]. There are two types of voting techniques: hard voting and soft voting. In hard voting, the final prediction is made by a majority vote in which the aggregator chooses the class prediction that appears repeatedly among the base models. On the other hand, the final class in soft-voting is the class with the highest probability averaged over the individual predictors [30]. In our work, we used hard voting, and the algorithm for hard voting is shown in Algorithm 1.

### 3.5. Proposed Model

We have proposed a robust ETSE model that outperforms previous studies. The entire procedure is presented in Algorithm 1. In this study, we applied seven ML models, including SVM, XGB, LGBM, DT, RF, ET, and K-Neighbors, to the training data. We applied Grid Search to each model to search for optimal parameter values. The optimal parameters obtained by Grid Search are presented in Table 3. By tuning the hyperparameters, we selected the most suitable model, which was then tested with the test dataset. We evaluated the model using the confusion matrix and calculated precision, recall, F1-score, and accuracy. The entire process is illustrated in Figure 3. In the final stage, we used a voting classifier to create an ensemble of two or three models based on their performance. We executed the ensemble of LGBM and SVM, XGB and SVM, Extra Trees and SVM, RF and SVM, DT and SVM, ET and LGBM, and analyzed the performance of each ensemble model. We found that the ensemble of Extra Tree and SVM was the most successful and proposed model. We also examined the ensemble of ET, SVM, and RF. The proposed ETSE model is illustrated in Figure 4.

**Algorithm 1: Proposed ETSE model Algorithm**

```
procedure Input(fetal_health_data)
return fetal_health_data [“Normal”, “Suspect”, “Pathological”]
procedure data_pre-processing
return processed_data
procedure Split_Data (fetal_health_data)
Training_data, Testing_data = split (fetal_health_data, label)
return Training_data, Testing-data
procedure M= [SVM, RF, DT, ET, XGB, LGBM, K-neighbor]
For n=1 to 7 do
    Parameters=Select the Hyperparameters to tune
    grid_result=GridSearchCV(M[n], Parameters)
    grid_result.best_params_
    M [n] =M [n] (grid_result.best_params_)
End
return tuned M= [SVM, RF, DT, ET, XGB, LGBM, K-neighbor]
For i=1 to 7 do
    M [i]=M[i] (Training_data, Training_label, Testing_data)
    Prediction=M[i](Testing_data)
    M [i]. evaluate (Testing_data)
End
voting = “hard”
M [4] = ExtraTrees (Training_data, Training_label, Testing_data)
M [1] = SVM (Training_data, Training_label, Testing_data)
procedure Ensemble_Model (Training_data, Training_label, Testing_data)
hard_voting_classifier= concatenate (M [1], M [4])
hard_voting_classifer.fit (Training_data, Training_label)
Predictions = hard.voting_classifier.predict (Testing_data)
```

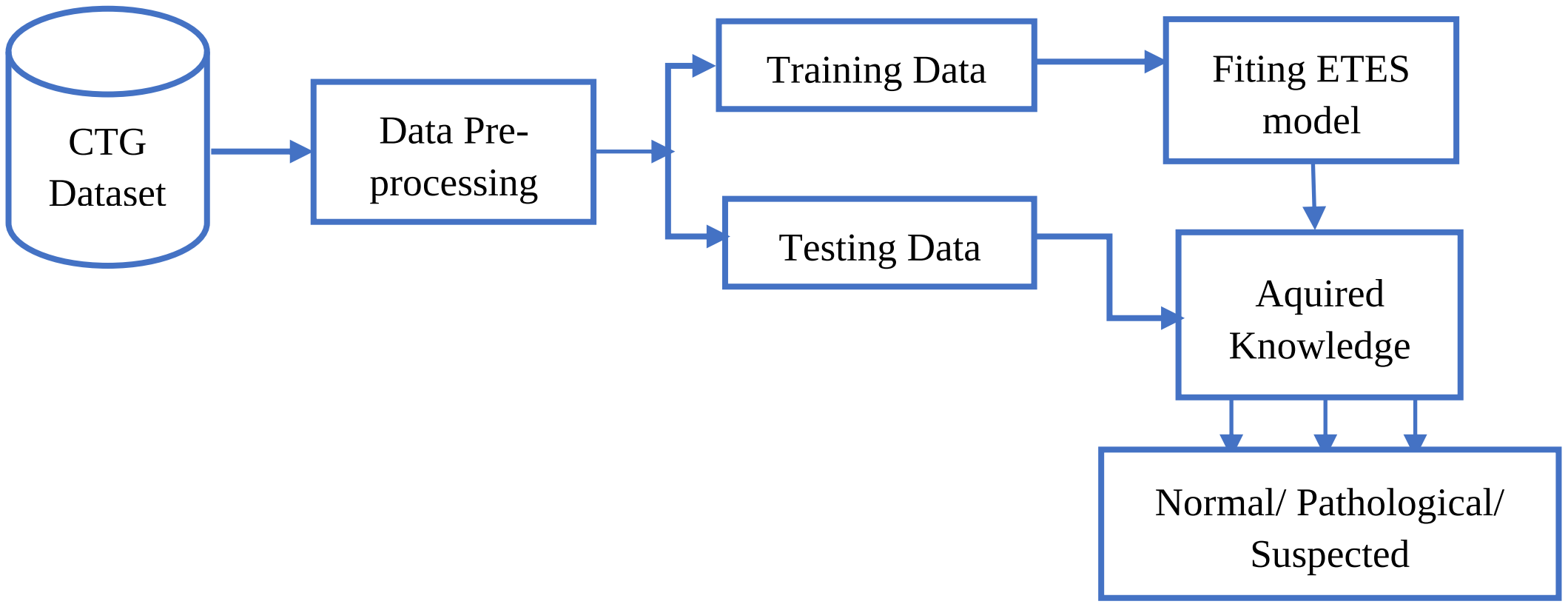

**Figure 3**. Proposed Methodology.

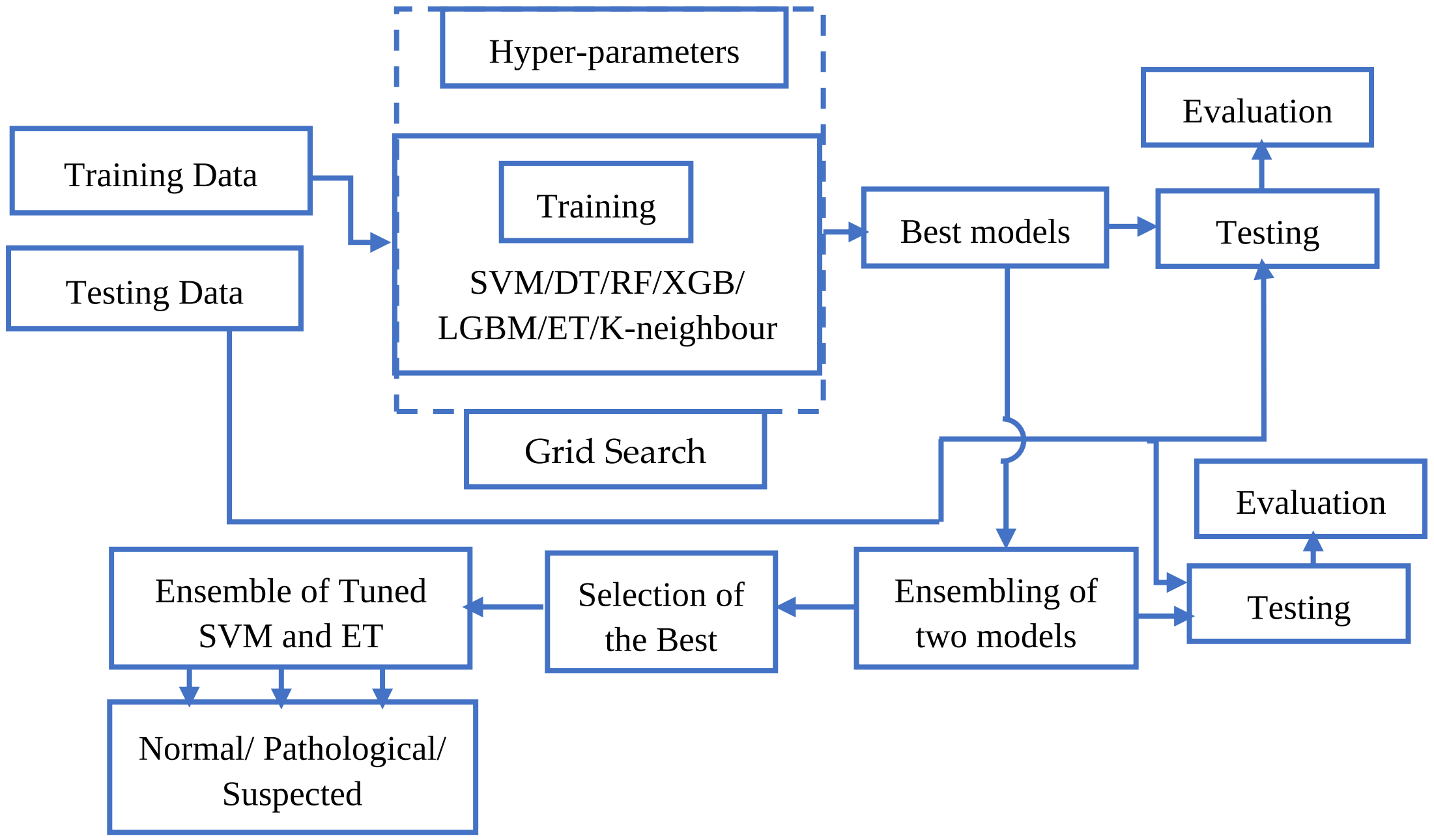

**Figure 4**. Proposed Ensemble of Tuned SVM and ET (ETSE) model.

**Table 3.** Tuned parameters.

| Algorithm | Parameters |
|---|---|
| SVM | C=1000, decision_function_shape='ovo', gamma=0.9, probability=True |
| XGB | max_depth=12, n_estimators=200 |
| LGBM | boosting_type='goss', max_depth=200, random_state=100, silent=True, metric='None', n_jobs=4, num_leaves=20, n_estimators=100 |
| DT | ccp_alpha=0.001, criterion='entropy', max_depth=12, max_features='log2' |
| RF | max_features='auto', n_estimators= 100, max_depth=12, criterion='entropy' |
| ET | random_state=1, n_estimators=340, max_features=None, max_depth=25, criterion='gini' |
| K-Neighbors | n_neighbors=3, p= 1, weights= 'distance' |

### 3.6. Model Performance Analysis

In order to evaluate the performance of our proposed model, we employ both quantitative and qualitative methods. Qualitative evaluation is a standard technique for visually comparing and evaluating image classification results [31]. Specifically, we assess our proposed model using a confusion matrix and a set of performance measures including accuracy, precision, recall, and F1 score. These measures are defined using the following equations:

$$\text{Accuracy} = \frac{\sum TP + \sum TN}{\sum TP + \sum TN + \sum FP + \sum FN} * 100 \quad (2)$$

$$\text{Precision} = \frac{\sum TP}{\sum TP + \sum FP} * 100 \quad (3)$$

$$\text{Recall} = \frac{\sum TP}{\sum TP + \sum FN} \quad (4)$$

$$\text{F1 score} = 2 * \left(\frac{Precision * Recall}{Precision + Recall}\right) * 100 \quad (5)$$

$$\text{Macro Avg Measure} = \frac{1}{N}(Measure\ in\ class_1 + Measure\ in\ class_2 + \cdots + Mesure\ in\ class_N) \quad (6)$$

$$\text{Weighted Avg Measure} = \frac{(Measure * weight)\ in\ Class_1 + (Measure * weight) in\ class_2 + \cdots + (Measure * weight) in\ class_N}{Total\ number\ of\ sample} \quad (7)$$

Where: TP denotes the true positive; TN is true negative; FP denotes the false positive; FN refers to the false negative. In macro-averaged, all classes equally contribute to the final averaged metric but in weighted-average, each class contribution to the average is weighted by its size.

## 4. Result Analysis

In our experiment, we successfully predicted fetal health using seven different machine learning models, namely SVM, XGB, KGBM, DT, RF, ExtraTrees, and K-Neighbor. Tuned parameters for each model were obtained through a grid search, and the confusion matrix for each model was presented in Figure 5. The number of misclassifications for the tuned SVM, XGB, KGBM, DT, RF, ExtraTrees, and K-Neighbor models are 9, 22, 22, 41, 22, and 52, respectively. In Table 4, the precision, recall, and F1 score of each model are calculated for each class. The K-neighbor classifier has poor precision for the suspect class and poor recall for the normal class. On the other hand, the SVM classifier shows 100% precision, 100% recall, and 100% F1 score for the pathological class and 99% precision, 99% recall, and 99% F1 score for the rest two classes. The other classifiers show precision, recall, and F1 score values between SVM and K-Neighbor. In Table 5, the overall precision, recall, F1 score, and accuracy of each model are calculated. The tuned SVM model yields the highest precision, recall, and F1 score of 99% and an accuracy of 99.39%, while K-Neighbor has the lowest precision, recall, F1 score, and accuracy of 97%, 97%, 96%, and 96.51%, respectively.

To improve the classification accuracy, we created ensemble models by combining the top-performing classifiers. We tested ten ensemble models, including LGBM+SVM, XGB+SVM,

ExtraTrees+SVM, RF+SVM, DT+SVM, ET+LGBM, DT+ET, XGB+ET, XGB+LGBM, and SVM+ET+RF. Figure 5 displays the confusion matrix for each ensemble model. The LGBM+SVM, XGB+SVM, RF+SVM, and RF+SVM ensemble models showed the most significant improvement in precision, recall, and F1 score, resulting in an accuracy of 99.53%, which is higher than SVM's accuracy of 99.39%. However, other ensemble models like DT+ SVM, ET+LGBM, DT+ET, XGB+ET, and XGB+LGBM did not outperform the tuned SVM. The proposed Ensemble of Tuned SVM and ExtraTrees (ETSE) model showed a significant improvement in precision, recall, and F1 score, achieving 100% precision, recall, and F1 score, and accuracy of 99.66%. The ensemble of the best three classifiers (SVM+ET+RF) could not improve the classification accuracy of this dataset. In conclusion, our proposed ETSE model has been proved as the best model for fetal health prediction.

**Table 4**. Performance Analysis of models for individual classes.

| **Model** | **Normal** | | | **Suspect** | | | **Pathological** | | |
|---|---|---|---|---|---|---|---|---|---|
| | **Precision (%)** | **Recall (%)** | **F1-score (%)** | **Precision (%)** | **Recall (%)** | **F1-score (%)** | **Precision (%)** | **Recall (%)** | **F1-score (%)** |
| SVM | 99 | 99 | 99 | 99 | 99 | 99 | 100 | 100 | 100 |
| XGB | 99 | 96 | 98 | 97 | 99 | 98 | 100 | 100 | 100 |
| LGBM | 99 | 96 | 98 | 97 | 99 | 98 | 100 | 100 | 100 |
| Decision Tree | 99 | 93 | 96 | 94 | 99 | 96 | 99 | 100 | 99 |
| Random Forest | 99 | 96 | 98 | 96 | 99 | 98 | 100 | 100 | 100 |
| Extra Trees | 99 | 97 | 98 | 97 | 99 | 98 | 100 | 100 | 100 |
| K-Neighbors | 99 | 90 | 95 | 91 | 99 | 95 | 100 | 100 | 100 |
| LGBM+SVM | 99 | 99 | 99 | 99 | 99 | 99 | 100 | 100 | 100 |
| XGB+SVM | 99 | 99 | 99 | 99 | 99 | 99 | 100 | 100 | 100 |
| RF+SVM | 99 | 99 | 99 | 99 | 99 | 99 | 100 | 100 | 100 |
| DT+ SVM | 99 | 99 | 99 | 97 | 99 | 98 | 100 | 98 | 99 |
| ET+LGBM | 99 | 98 | 99 | 98 | 99 | 99 | 100 | 100 | 100 |
| DT+ET | 98 | 98 | 98 | 98 | 98 | 98 | 100 | 100 | 100 |
| XGB+ET | 99 | 98 | 99 | 98 | 99 | 99 | 100 | 100 | 100 |
| XGB+LGBM | 99 | 97 | 98 | 97 | 99 | 98 | 100 | 100 | 100 |
| SVM+ET+RF | 99 | 97 | 98 | 97 | 99 | 98 | 100 | 100 | 100 |
| **ETSE model (Extra Trees + SVM)** | **99** | **100** | **99** | **100** | **99** | **99** | **100** | **100** | **100** |

**Table 5**. Overall Performance Analysis of models.

| Model | Precision | Recall | F1-score | Accuracy |
|---|---|---|---|---|
| SVM | 99 | 99 | 99 | 99.39 |
| XGB | 99 | 99 | 99 | 98.52 |
| LGBM | 99 | 99 | 99 | 98.52 |
| Decision Tree | 97 | 97 | 97 | 97.24 |
| Random Forest | 98 | 98 | 98 | 98.45 |
| Extra Trees | 99 | 99 | 99 | 98.79 |
| KNeighbors | 97 | 97 | 96 | 96.51 |
| LGBM+SVM | 100 | 100 | 100 | 99.53 |
| XGB+SVM | 100 | 100 | 100 | 99.53 |
| RF+SVM | 100 | 100 | 100 | 99.53 |
| DT+ SVM | 99 | 99 | 99 | 98.66 |
| ET+LGBM | 99 | 99 | 99 | 99.06 |
| DT+ET | 99 | 99 | 99 | 98.52 |
| XGB+ET | 99 | 99 | 99 | 99.06 |
| XGB+LGBM | 99 | 99 | 99 | 98.86 |
| SVM+ET+RF | 99 | 99 | 99 | 98.86 |
| **ETSE model (Extra Trees + SVM)** | **100** | **100** | **100** | **99.66** |

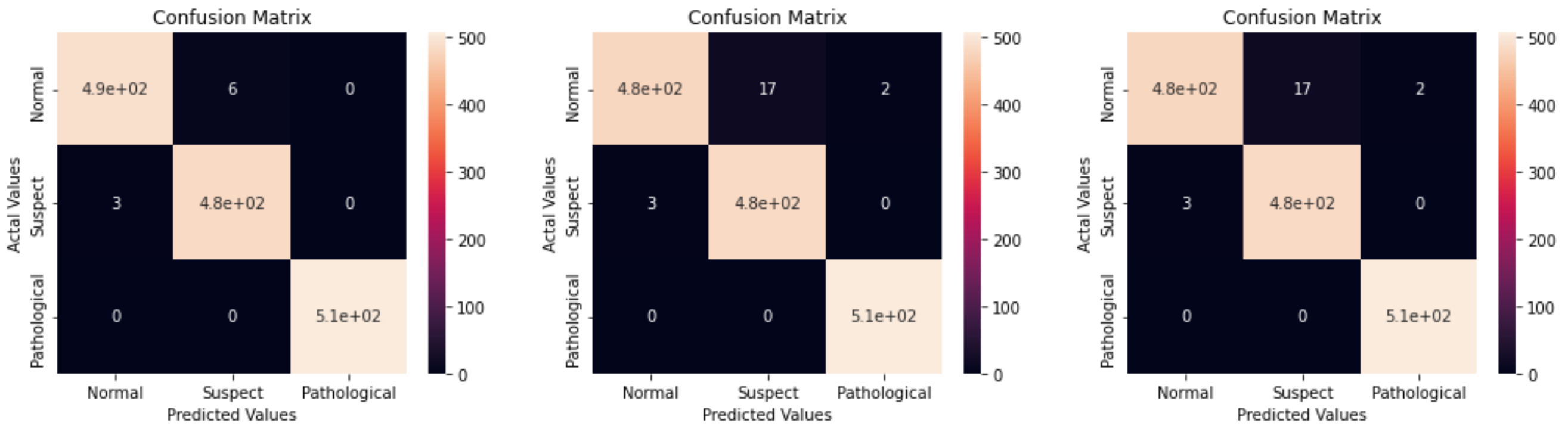

(a) SVM Classifier (b) XGB Classifier (c) LGBM Classifier

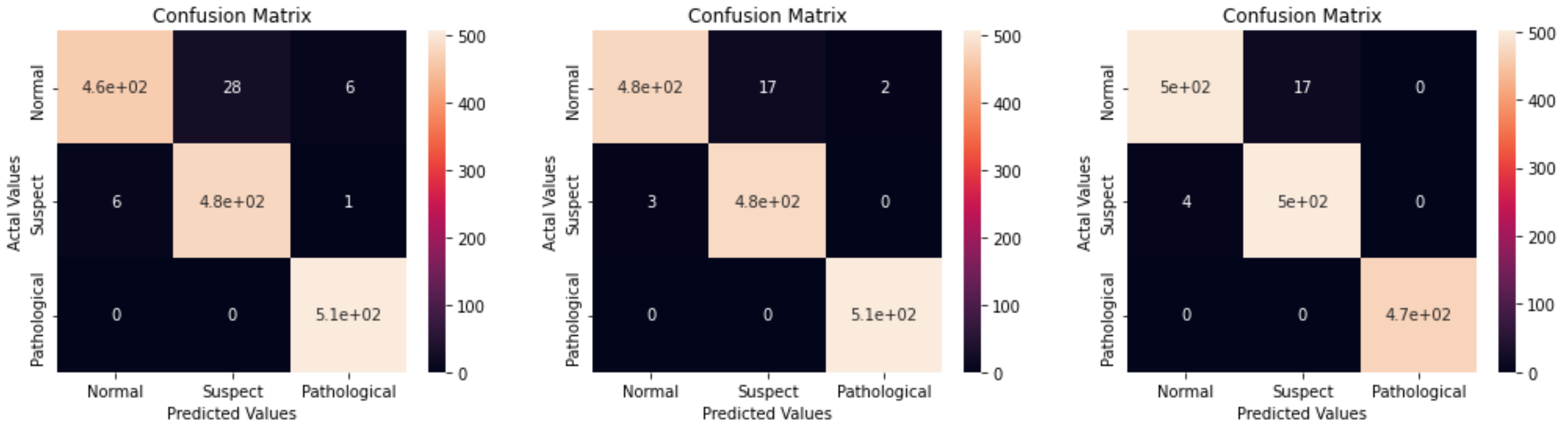

(d) DT Classifier (e) RF Classifier (f) ET Classifier

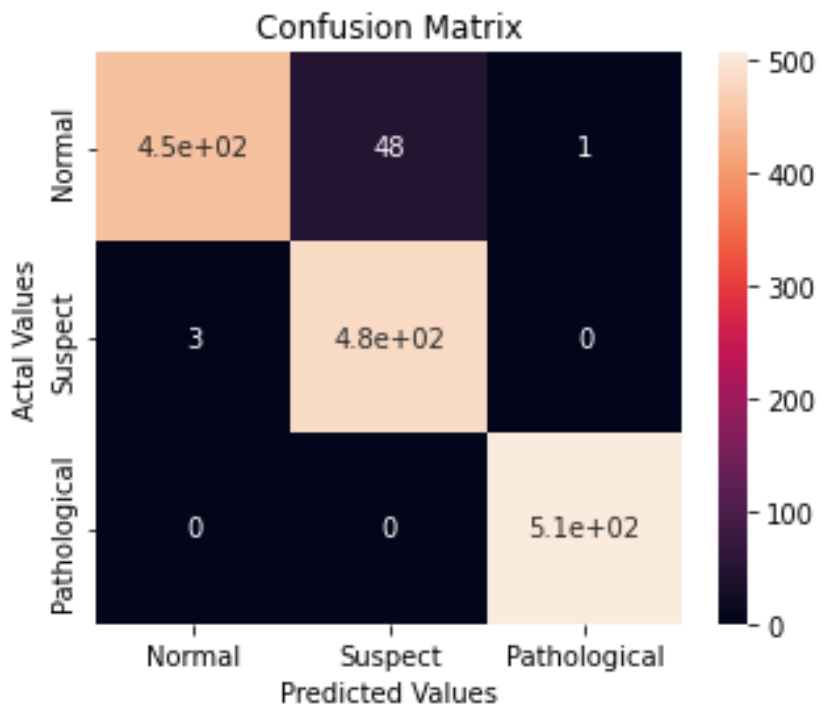

(g) K- Neighbours Classifier

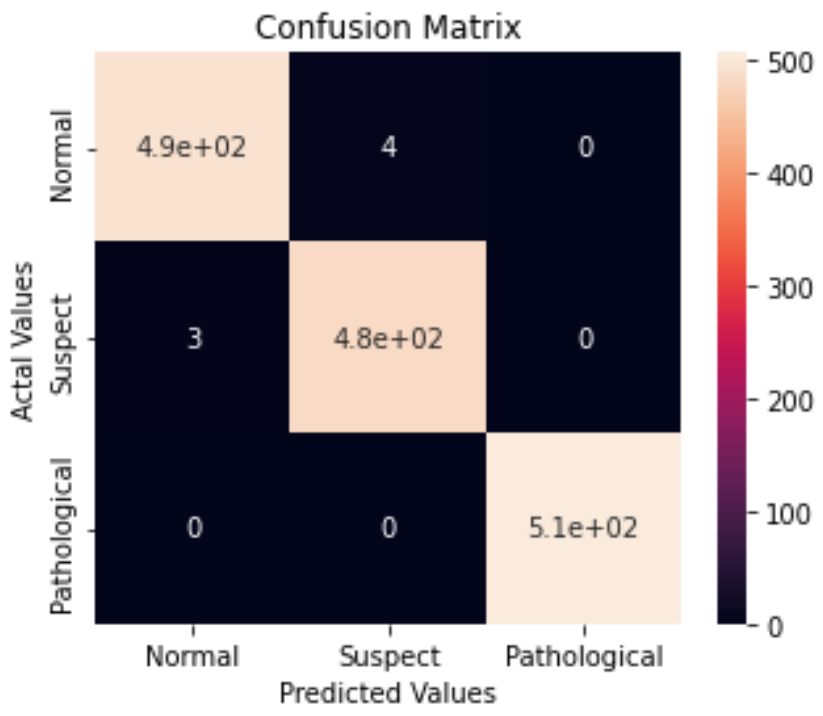

(h) LGBM+SVM Classifier

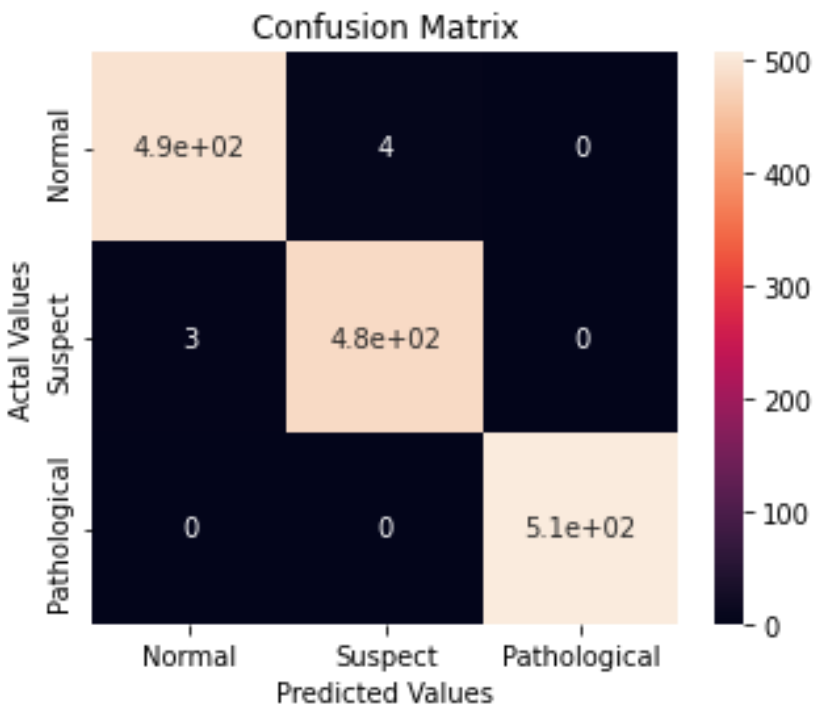

(i) XGB+SVM Classifier

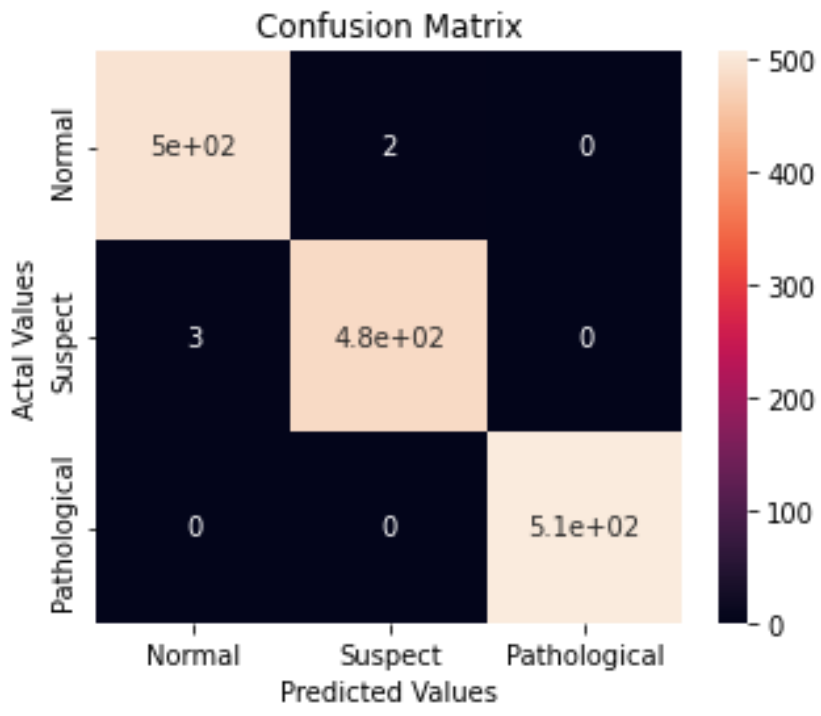

(j) **ETSE (ET+SVM) Classifier**

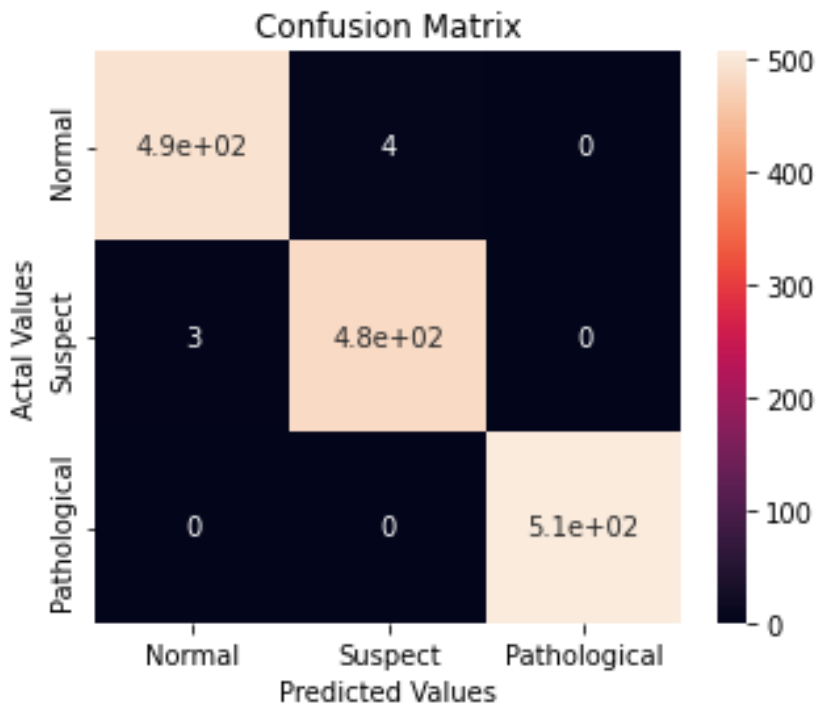

(k) RF+SVM Classifier

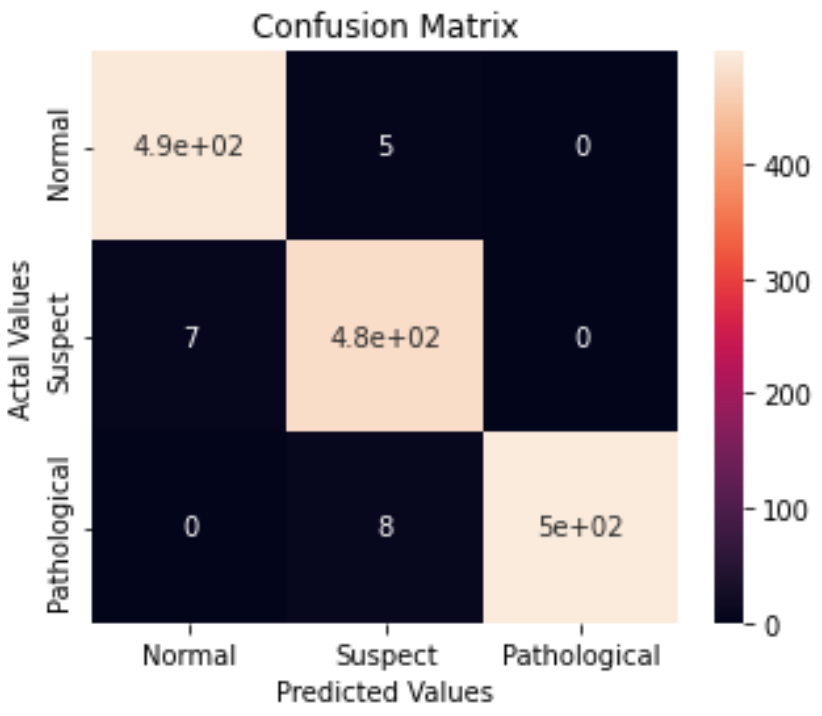

(l) DT+SVM Classifier

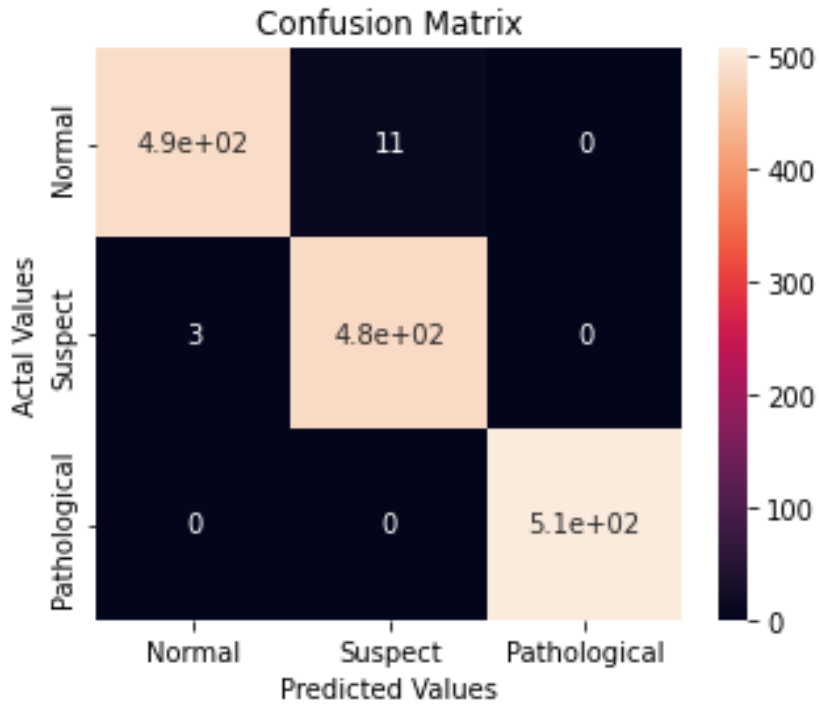

(m) ET+LGBM Classifier

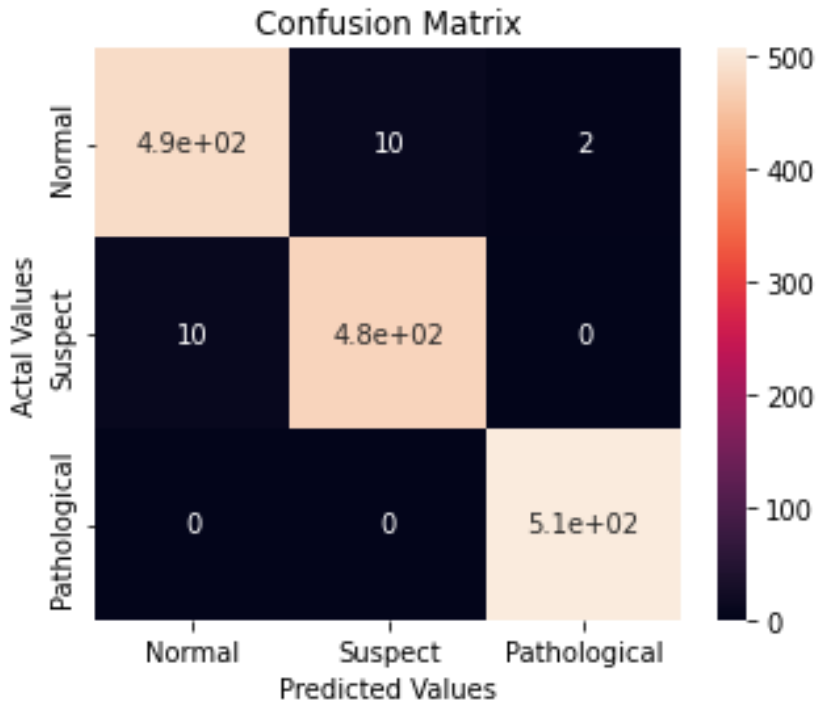

(n) DT+ET Classifier

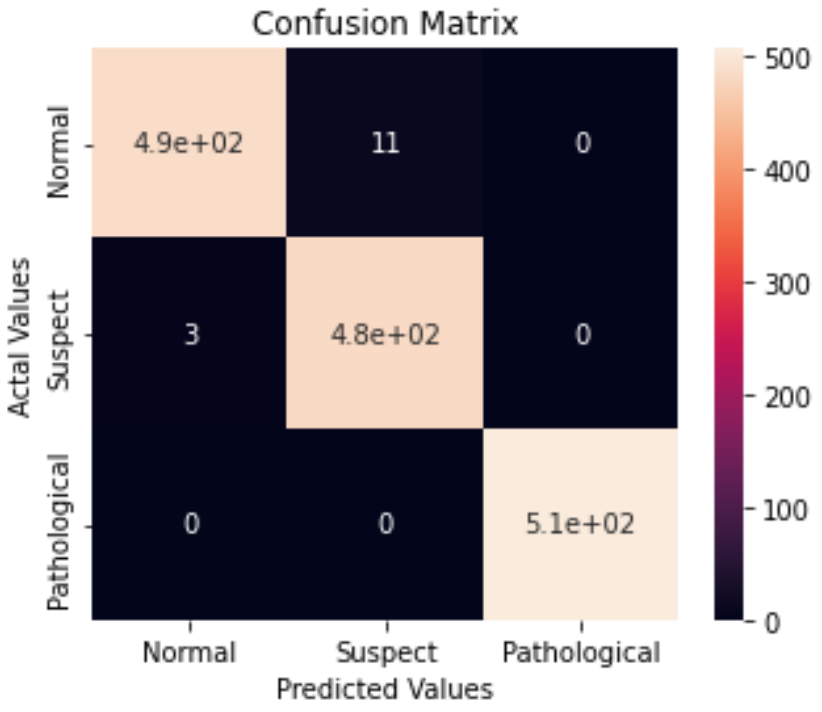

(0)XGB+ET Classifier

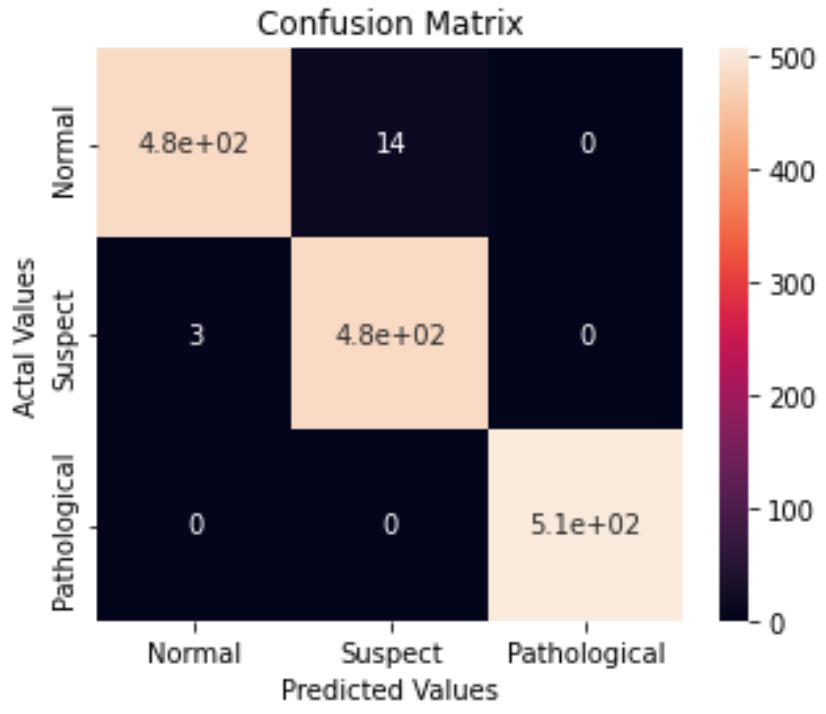

(p) XGB+LGBM Classifier

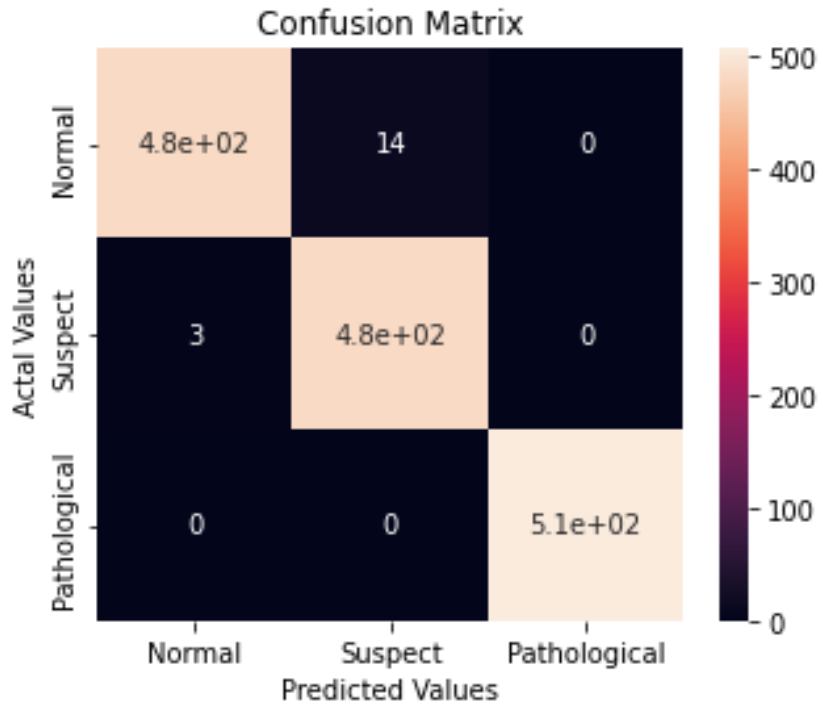

(q) ET+SVM+RF Classifier

**Figure 5**. Confusion Matrix of Each model.

## 5. Discussion and Conclusion

The performance of individual models (SVM, XGB, LGBM, DT, RF, ET, and K-Neighbors) has significantly improved compared to the previous study [20], as shown in Table 6. Two main reasons account for the increased accuracy of the models: first, data preprocessing operations such as outlier rejection, missing value imputation, data standardization, and data sampling; second, hyperparameter tuning of the models using GridSearchCV. The SVM classifier achieved the highest accuracy of 99.39%, with 99% precision, recall, and F1-score. Furthermore, ensemble of different models was investigated to further enhance accuracy. The ET and SVM ensemble achieved the highest accuracy of 99.66%, with 100% precision, recall, and F1-score. The application of ensemble learning and hyperparameter tuning, which were not used in the reference [20], played a critical role in achieving these results. This proposed ETSE model can be a valuable tool in detecting fetal health in the womb and reducing fetal deaths, resulting in a healthier infant.

**Table 6.** Comparisons of our work.

| Model | Precision | Recall | F1-score | Accuracy |
|---|---|---|---|---|
| SVM [20] | 97 | 97 | 97 | 96.57 |
| RF [20] | 98 | 98 | 98 | 97.51 |
| DT [20] | 96 | 96 | 96 | 95.70 |
| K-Neighobour [20] | 91 | 90 | 90 | 90.20 |
| SVM (our paper) | 99 | 99 | 99 | 99.39 |
| RF (our paper) | 98 | 98 | 98 | 98.45 |
| DT (our paper) | 97 | 97 | 97 | 97.24 |
| K-Neighobour (our paper) | 97 | 97 | 96 | 96.51 |
| **Our proposed ETSE (ET + SVM)** | **100** | **100** | **100** | **99.66** |

This work has a few limitations that should be acknowledged. Firstly, the model has not been tested in real-time clinical settings, which may limit its generalizability and practical application. Secondly, due to the unavailability of different datasets, the model has not been validated on other datasets, which may affect its robustness and reliability. Another limitation is the scarcity of references in this field, which may affect the comprehensiveness of the study. Finally, the deployment of the model is not discussed in our work. There may be additional technical and logistical challenges that could arise in practical deployment. To address these limitations, future work could involve extensive validation of the model on patients at nearby hospitals and clinics.

Additionally, introducing new machine learning concepts such as weighted ensembles or other advanced techniques may further enhance the model's accuracy and performance.

**Author Contributions:** Md. Simul Hasan Talukder implemented the model in Google-Colab with GPU. Sharmin Akter suggested machine learning concept and contributed in manuscripts writing, editing and correcting.

**Funding:** This research received no external funding.

**Conflicts of Interest:** The authors declare no conflict of interest.

## References

1. O'Neill E, Thorp J. Antepartum evaluation of the fetus and fetal well being. Clin Obstet Gynecol. 2012 Sep;55(3):722-30. doi: 10.1097/GRF.0b013e318253b318. PMID: 22828105; PMCID: PMC3684248.
2. H. Ocak and H. Ertunc, "Prediction of fetal state from the cardiotocogram recordings using adaptive neuro-fuzzy inference systems", Neural Computing and Applications, vol. 23, no. 6, pp. 1583-1589, 2012. Available: 10.1007/s00521-012-1110-3 [Accessed 2 July 2022].
3. Sundar, C. (2012). An Analysis on the Performance of K-Means Clustering Algorithm for Cardiotocogram Data Clustering. International Journal on Computational Science & Applications, 2(5), 11–20. https://doi.org/10.5121/ijcsa.2012.2502
4. Arif, M. (2015). Classification of cardiotocograms using random forest classifier and selection of important features from cardiotocogram signal. Biomaterials and Biomechanics in Bioengineering, 2(3), 173–183. https://doi.org/10.12989/bme.2015.2.3.173.
5. I. Ingemarsson, "Fetal monitoring during labor," Neonatology, vol. 95, no. 4, pp. 342–346, 2009.
6. F. I. G. O. News, "Report of the FIGO study group on the assessment of new Technology," International Journal of Gynecology & Obstetrics, vol. 59, no. 2, pp. 169–173, 1997.
7. G. J. Miao, K. H. Miao, and J. H. Miao, "Neural pattern recognition model for breast cancer diagnosis," Multidisciplinary Journals in Science and Technology, Journal of Selected Areas in Bioinformatics, 2012.
8. K. H. Miao, J. H. Miao, and G. J. Miao, "Diagnosing coronary heart disease using ensemble machine learning," International Journal of Advanced Computer Science and Applications, vol. 7, no. 10, pp. 30–39, 2016.

9. K. H. Miao and G. J. Miao, “Mammographic diagnosis for breast cancer biopsy predictions using neural network classification model and receiver operating characteristic (ROC) curve evaluation,” Multidisciplinary Journals in Science and Technology, Journal of Selected Areas in Bioinformatics,vol. 3, no. 9, pp. 1–10, 2013.
10. J. H. Miao, K. H. Miao, and G. J. Miao, “Breast cancer biopsy predictions based on mammographic diagnosis using support vector machine learning,” Multidisciplinary Journals in Science and Technology, Journal of Selected Areas in Bioinformatics, Vol. 5, no. 4, pp. 1–9, 2015.
11. D. Ayres-de-Campos, J. Bernardes, A. Garrido, J. Marques-de-Sá, and L. Pereira-Leite, “Sisporto 2.0: a program for automated analysis of cardiotocograms,” The Journal of Maternal-Fetal Medicine, vol. 9, no. 5, pp. 311–318, 2000.
12. Hossen, M. D. A., Tazin, T., Khan, S., Alam, E., Sojib, H. A., Monirujjaman Khan, M., & Alsufyani, A. (2021). Supervised Machine Learning-Based Cardiovascular Disease Analysis and Prediction. Mathematical Problems in Engineering, 2021, 1–10. https://doi.org/10.1155/2021/1792201.
13. Saboor, A., Usman, M., Ali, S., Samad, A., Abrar, M. F., & Ullah, N. (2022). A Method for Improving Prediction of Human Heart Disease Using Machine Learning Algorithms. Mobile Information Systems, 2022, 1–9. https://doi.org/10.1155/2022/1410169
14. C, Sundar., M.Chitradevi, M. Chitradevi., & Geetharamani, G. (2012). Classification of Cardiotocogram Data using Neural Network based Machine Learning Technique. International Journal of Computer Applications, 47(14), 19–25. https://doi.org/10.5120/7256-0279
15. H. Ocak and H. M. Ertunc, “Prediction of fetal state from the cardiotocogram recordings using adaptive neuro-fuzzy inference systems,” Neural Computing Applications, vol. 23, no. 6,pp. 1583–1589, 2013.
16. H. Ocak, “A medical decision support system based on support vector machines and the genetic algorithm for the evaluation of fetal well-being,” Journal Medical System, vol. 37, no. 2, pp. 1–9, 2013.
17. Peterek, T., Gajdoš, P., Dohnálek, P., & Krohová, J. (2014). Human Fetus Health Classification on Cardiotocographic Data Using Random Forests. Advances in Intelligent Systems and Computing, 189–198. https://doi.org/10.1007/978-3-319-07773-4_19.

18. Mehbodniya, A., Lazar, A. J. P., Webber, J., Sharma, D. K., Jayagopalan, S., K, K., Singh, P., Rajan, R., Pandya, S., & Sengan, S. (2021). Fetal health classification from cardiotocographic data using machine learning. Expert Systems. https://doi.org/10.1111/exsy.12899
19. Rahmayanti, N., Pradani, H., Pahlawan, M., & Vinarti, R. (2022). Comparison of machine learning algorithms to classify fetal health using cardiotocogram data. Procedia Computer Science, 197, 162–171. https://doi.org/10.1016/j.procs.2021.12.130
20. Alam, M. T., Khan, M. A. I., Dola, N. N., Tazin, T., Khan, M. M., Albraikan, A. A., & Almalki, F. A. (2022). Comparative Analysis of Different Efficient Machine Learning Methods for Fetal Health Classification. Applied Bionics and Biomechanics, 2022, 1–12. https://doi.org/10.1155/2022/6321884
21. The UCI, "Machine learning repository, cardiotocography dataset," https://archive.ics.uci.edu/ml/datasets/cardiotocography .
22. N. V. Chawla, K. W. Bowyer, L. O. Hall, and W. P. Kegelmeyer, "SMOTE: synthetic minority over sampling technique," The Journal of Artificial Intelligence Research, vol. 16, pp. 321–357, 2002.
23. Cao, X. H., I. Stojkovic, and Z. Obradovic. (2016) "A robust data scaling algorithm to improve classification accuracies in biomedical data." BMC Bioinformatics 17 (1): 1-10.
24. Hyperparameter tuning. Grid search and random search. Available online: https://www.yourdatateacher.com/2021/05/19/hyperparameter-tuning-grid-search-and-random-search/#:~:text=Grid%20search%20is%20the%20simplest,performance%20metrics%20using%20cross%2Dvalidation.
25. Belete, D. M., & Huchaiah, M. D. (2021). Grid search in hyperparameter optimization of machine learning models for prediction of HIV/AIDS test results. International Journal of Computers and Applications, 1–12. https://doi.org/10.1080/1206212x.2021.1974663
26. Pedregosa F, Varoquaux G, Gramfort A, et al. Scikit-learn: machine learning in python. J Mach Learn Res. 2011; 12:2825–2830.
27. Bergstra J, Bengio Y. Random search for hyper-parameter optimization. J Mach Learn Res. 2012;13(1):281–305.

28. Akhter, M. P., Zheng, J., Afzal, F., Lin, H., Riaz, S., & Mehmood, A. (2021). Supervised ensemble learning methods towards automatically filtering Urdu fake news within social media. PeerJ Computer Science, 7, e425. https://doi.org/10.7717/peerj-cs.425
29. An ensemble approach for classification and prediction of diabetes mellitus using soft voting classifier. (2021). International Journal of Cognitive Computing in Engineering, 2, 40–46. https://doi.org/10.1016/j.ijcce.2021.01.001
30. González S, García S, Del Ser J, Rokach L, Herrera F. 2020. A practical tutorial on bagging and boosting based ensembles for machine learning: Algorithms, software tools, performance study, practical perspectives and opportunities. Information Fusion 64(1):205–237.DOI 10.1016/j.inffus.2020.07.007.
31. M. F. Taha et al., "Using Deep Convolutional Neural Network for Image-Based Diagnosis of Nutrient Deficiencies in Plants Grown in Aquaponics," *Chemosensors*, 2022, Volume 10, p. 45, doi: 10.3390/chemosensors10020045.
32. Naveen Reddy Navuluri, 2021, Fetal Health Prediction using Classification Techniques, INTERNATIONAL JOURNAL OF ENGINEERING RESEARCH & TECHNOLOGY (IJERT) Volume 10, Issue 11 (November 2021).
33. Hasan, Md. K., Alam, Md. A., Das, D., Hossain, E., & Hasan, M. (2020). Diabetes Prediction Using Ensembling of Different Machine Learning Classifiers. IEEE Access, 8, 76516–76531. https://doi.org/10.1109/access.2020.2989857
34. Mohan, S., Thirumalai, C., & Srivastava, G. (2019). Effective Heart Disease Prediction Using Hybrid Machine Learning Techniques. IEEE Access, 7, 81542–81554. https://doi.org/10.1109/access.2019.2923707
35. Kuhle, S., Maguire, B., Zhang, H., Hamilton, D., Allen, A. C., Joseph, K. S., & Allen, V. M. (2018). Comparison of logistic regression with machine learning methods for the prediction of fetal growth abnormalities: a retrospective cohort study. BMC Pregnancy and Childbirth, 18(1). https://doi.org/10.1186/s12884-018-1971-2
36. Muntasir Nishat, M., Faisal, F., Jahan Ratul, I., Al-Monsur, A., Ar-Rafi, A. M., Nasrullah, S. M., Reza, M. T., & Khan, M. R. H. (2022). A Comprehensive Investigation of the Performances of Different Machine Learning Classifiers with SMOTE-ENN Oversampling Technique and Hyperparameter Optimization for Imbalanced Heart Failure Dataset. Scientific Programming, 2022, 1–17. https://doi.org/10.1155/2022/3649406

37. Karthick, K., Aruna, S. K., Samikannu, R., Kuppusamy, R., Teekaraman, Y., & Thelkar, A. R. (2022). Implementation of a Heart Disease Risk Prediction Model Using Machine Learning. Computational and Mathematical Methods in Medicine, 2022, 1–14. https://doi.org/10.1155/2022/6517716
38. Mahesh, T. R., Dhilip Kumar, V., Vinoth Kumar, V., Asghar, J., Geman, O., Arulkumaran, G., & Arun, N. (2022). AdaBoost Ensemble Methods Using K-Fold Cross Validation for Survivability with the Early Detection of Heart Disease. Computational Intelligence and Neuroscience, 2022, 1–11. https://doi.org/10.1155/2022/9005278
39. Itoo, F., & Singh, S. (2021). Comparison and analysis of logistic regression, Naïve Bayes and KNN machine learning algorithms for credit card fraud detection. International Journal of Information Technology, 13, 1503-1511.
40. Sarwar, A., Ali, M., Manhas, J., & Sharma, V. (2020). Diagnosis of diabetes type-II using hybrid machine learning based ensemble model. International Journal of Information Technology, 12, 419-428.
41. Mittal, K., Aggarwal, G., & Mahajan, P. (2019). Performance study of K-nearest neighbor classifier and K-means clustering for predicting the diagnostic accuracy. International Journal of Information Technology, 11, 535-540.
42. Singh, K., Kumar, S., & Kaur, P. (2019). Support vector machine classifier based detection of fungal rust disease in Pea Plant (Pisam sativam). International Journal of Information Technology, 11, 485-492.
43. Talukder, M. S. H., & Sarkar, A. K. (2023). Nutrients deficiency diagnosis of rice crop by weighted average ensemble learning. Smart Agricultural Technology, 4, 100155.